\documentclass{article}

\PassOptionsToPackage{numbers, compress}{natbib}

 \usepackage[preprint]{neurips_2026}

\usepackage[utf8]{inputenc} 
\usepackage[T1]{fontenc}    
\usepackage{hyperref}       
\usepackage{url}            
\usepackage{booktabs}       
\usepackage{amsfonts}       
\usepackage{nicefrac}       
\usepackage{microtype}      
\usepackage{xcolor}         

\usepackage{xspace}

\usepackage[most]{tcolorbox}
\usepackage{fvextra}

\usepackage{hyperref}
\usepackage{url}
\usepackage{amsmath}

\usepackage{graphicx}
\usepackage{subcaption}
\usepackage{float}

\usepackage{pgfplots}
\usepgfplotslibrary{groupplots}
\pgfplotsset{compat=1.18}

\definecolor{cecolor}{RGB}{140,81,10}     
\definecolor{ntcecolor}{RGB}{46,139,87}   
\definecolor{sntcecolor}{RGB}{255,191,0}  
\definecolor{nonlcolor}{RGB}{65,105,225}  

\definecolor{cifar10best}{RGB}{0, 120, 0}      
\definecolor{cifar100best}{RGB}{200, 100, 0}   
\definecolor{baseline}{RGB}{248, 248, 248}     
\definecolor{header}{RGB}{250, 250, 250}       

\definecolor{bestresult}{HTML}{228B22}  
\definecolor{HeaderGray}{RGB}{248,249,250}
\definecolor{AccentGray}{RGB}{242,244,246}
\definecolor{SubtleHighlight}{RGB}{250,251,252}
\definecolor{TextGray}{RGB}{90,90,90}
\definecolor{BorderLight}{RGB}{220,220,220}
\definecolor{MethodGray}{RGB}{110,110,110}

\usepackage{array}
\usepackage{booktabs,multirow,siunitx,xcolor,colortbl}
\definecolor{header}{RGB}{245,245,245}

\definecolor{lightgray}{gray}{0.95}
\definecolor{darkgray}{gray}{0.8}

\usepackage{tikz}
\usepackage{pgfplots}
\usepackage{subcaption}
\usepackage{xcolor}

\usetikzlibrary{arrows.meta, positioning, shapes}
\usepackage{amssymb}
\usepackage{amsmath,amsfonts}
\usepackage{amsthm}
\usepackage{array}
\usepackage{booktabs}

\usepackage{array}
\usepackage{makecell}

\usepackage[table,xcdraw,dvipsnames]{xcolor}
\usepackage{colortbl}

\usepackage{graphicx}
\usepackage{tikz}
\usepackage{pgfplots}
\usepackage{subcaption}

\usepackage{adjustbox}
\usepackage{multirow}
\usepackage{multicol}
\usepackage{makecell}
\usepackage{tabularx}
\usepackage{float}
\usepackage{placeins}
\usepackage{stfloats}
\usepackage{lscape}
\usepackage{afterpage}

\usepackage{bm}
\usepackage{algorithmic}
\usepackage{fontawesome5}
\usepackage{pifont}
\usepackage{textcomp}

\usepackage{enumitem}

\usepackage{xspace}
\usepackage{url}
\usepackage{verbatim}

\usepackage{cite}
\usepackage{xr}

\usepackage{cleveref}

\newcommand{\eg}{\emph{e.g.}\xspace}
\newcommand{\ie}{\emph{i.e.}\xspace}

\definecolor{ForestGreen}{RGB}{34, 139, 34}
\definecolor{BrickRed}{RGB}{178, 34, 34}
\definecolor{NavyBlue}{RGB}{0, 100, 200}

\definecolor{lim1color}{HTML}{2C5282}  
\definecolor{lim2color}{HTML}{9C4221}  

\newcommand{\Lone}{\textcolor{lim1color}{\textbf{L1}}\xspace}
\newcommand{\Ltwo}{\textcolor{lim2color}{\textbf{L2}}\xspace}

\newcommand{\limone}[1]{\textcolor{lim1color}{\textbf{Limitation 1 (L1): #1}}}
\newcommand{\limtwo}[1]{\textcolor{lim2color}{\textbf{Limitation 2 (L2): #1}}}

\def\ie{\textit{i.e.}\xspace}
\def\eg{\textit{e.g.}\xspace}

\DeclareRobustCommand{\acronym}{\texttt{fmxcoders}\xspace}
\DeclareRobustCommand{\acronymsingle}{\texttt{fmxcoder}\xspace}

\usepackage{amsmath,amsfonts,bm}

\def\eqref#1{equation~\ref{#1}}

\def\1{\bm{1}}

\def\va{{\bm{a}}}
\def\vb{{\bm{b}}}
\def\vc{{\bm{c}}}
\def\vd{{\bm{d}}}
\def\ve{{\bm{e}}}

\def\vm{{\bm{m}}}

\def\vu{{\bm{u}}}
\def\vv{{\bm{v}}}
\def\vw{{\bm{w}}}
\def\vx{{\bm{x}}}

\def\vz{{\bm{z}}}

\def\mD{{\bm{D}}}
\def\mE{{\bm{E}}}

\def\mM{{\bm{M}}}
\def\mN{{\bm{N}}}

\def\mS{{\bm{S}}}

\def\mU{{\bm{U}}}
\def\mV{{\bm{V}}}
\def\mW{{\bm{W}}}
\def\mX{{\bm{X}}}
\def\mY{{\bm{Y}}}

\DeclareMathAlphabet{\mathsfit}{\encodingdefault}{\sfdefault}{m}{sl}
\SetMathAlphabet{\mathsfit}{bold}{\encodingdefault}{\sfdefault}{bx}{n}
\newcommand{\tens}[1]{\bm{\mathsfit{#1}}}

\def\tD{{\tens{D}}}
\def\tE{{\tens{E}}}

\def\tG{{\tens{G}}}

\def\tU{{\tens{U}}}

\def\gO{{\mathcal{O}}}

\def\sR{{\mathbb{R}}}

\newcommand{\cS}{\mathcal{S}}

\DeclareMathOperator{\Tr}{Tr}

\title{\acronym: Factorized Masked Crosscoders for Cross-Layer Feature Discovery}

\author{%
  Andreas D. Demou$^{1}$\quad
  Panagiotis Koromilas$^{1,2}$\quad
  James Oldfield$^{3}$ \\[2pt]
  \bfseries
  Yannis Panagakis$^{2,4}$\quad
  Mihalis A. Nicolaou$^{5,1}$ \\[6pt]
  \normalfont
  $^{1}$The Cyprus Institute\quad
  $^{2}$University of Athens\quad
  $^{3}$University of Oxford \\
  $^{4}$Archimedes AI/Athena Research Center\quad
  $^{5}$University of Cyprus
}

\begin{document}

\maketitle

\begin{abstract}
\looseness-1Many features in pretrained Transformers span multiple layers: they emerge through stages of inference, persist in the residual stream, or are built jointly by parallel MLPs. Crosscoders (namely, sparse dictionaries trained jointly across layers) aim to recover these cross-layer features in a single shared latent space. We show that standard crosscoders largely fail at this purpose. Although their decoder weight norms spread evenly across layers, a functional coherence metric we introduce reveals that each latent's activation is effectively driven by only one or two layers on average. While functionally coherent latents act as human-interpretable concept detectors (e.g., US states and cities), the layer-localized latents that crosscoders predominantly learn collapse onto surface-level patterns such as digit detectors. We trace this failure to two structural limitations: unconstrained cross-layer parameterization and unregularized cross-layer dependence. We address both by introducing \acronym, which (i) replace the encoder and decoder with low-rank tensor factorizations that draw every latent's per-layer weights from a shared cross-layer basis, and (ii) apply stochastic layer masking, a denoising regularizer along the layer axis that penalizes latents whose contribution collapses when a single layer is masked. Across GPT2-Small, Pythia-410M, Pythia-1.4B, and Gemma2-2B, \acronym lift mean probing F1 by 10–30 points, surpassing per-layer SAE baselines that standard crosscoders fail to reach, reduce reconstruction MSE by 25–50\%, and roughly double mean functional coherence. An LLM-as-a-judge evaluation further shows that \acronym recover 3–13$\times$ more semantically coherent latents than standard crosscoders across all four base LLMs.
\end{abstract}

\section{Introduction}\label{sec:intro}
\begin{figure}[ht!]
    \centering
    \includegraphics[width=\textwidth]{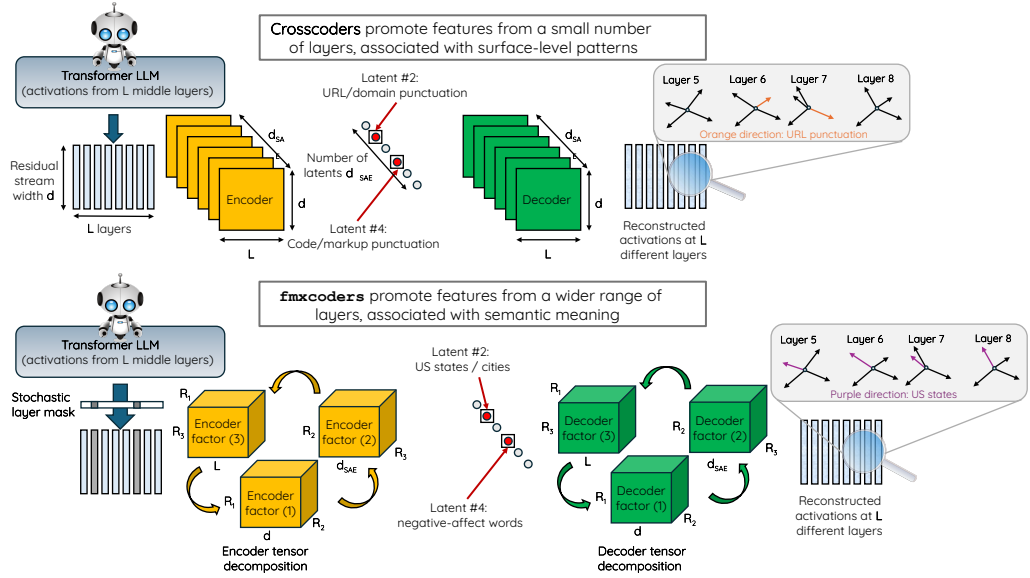}
    \caption{Standard crosscoders (\textbf{top}) parameterize the encoder and decoder as $L$ independent per-layer matrices and recover latents whose functional dependence collapses onto a few layers, capturing surface-level patterns. \acronym (\textbf{bottom}) replace these with low-rank tensor factorizations that share a cross-layer basis across all latents, and add stochastic layer masking as a denoising regularizer along the layer axis, producing latents that span more layers and act as semantic concept detectors. Insets show per-layer decoder directions for one example latent.}
    \label{fig:overview}
\end{figure}
The principal aim of mechanistic interpretability is the decomposition of network activations into human-interpretable features and the identification of circuits that connect these features \citep{elhage2022toymodels, sharkey2025openproblems}. Focusing on the former, the dominant method for feature detection is sparse dictionary learning, which trains an encoder to map activations into a sparse code and a decoder to reconstruct them \citep{bricken2023towards, cunningham2023sparse, templeton2024scaling, gao2024topk, rajamanoharan2024jumprelu, bussmann2024batchtopk, koromilas2026polysae}. Each coordinate of the sparse code is referred to as a \emph{latent}, and corresponds to a learned dictionary direction whose activation is interpreted as the presence of a feature in the input.

A deep network gives rise to two kinds of features that dictionary learning aims to recover: \emph{local} features, tied to a specific layer or a narrow band of adjacent layers, and \emph{global} features, built jointly by several layers or persisting across many. Both are well documented in the literature. On the local side, knowledge neurons store specific factual associations at identifiable layers~\citep{dai2022knowledge}. At a coarser scale, transformer inference appears to unfold in universal stages (detokenization, feature engineering, prediction ensembling, and residual sharpening) each tied to a range of layers~\citep{lad2025remarkable}. On the global side, the residual-stream view holds that transformer layers incrementally update a shared hidden state across the network~\citep{geva2022transformer}. Universal neurons recur with similar activation patterns across different LLMs~\citep{gurnee2024universal}. Per-layer sparse autoencoder (SAE) analyses reveal features that activate with near-identical directions across many consecutive layers~\citep{lawson2025mlsae, balagansky2025permutability}, as well as features built by multiple MLPs acting in parallel on the residual stream, forming \emph{cross-layer superposition}~\citep{lindsey2024crosscoders}.

SAEs are able to properly recover local features \citep{bricken2023towards, cunningham2023sparse} by tying a dictionary to a single layer. Crosscoders \citep{lindsey2024crosscoders, ameisen2025circuit} extend this to attain global features by learning a single dictionary whose latent neurons read from and write to multiple layers at once. Although crosscoder weight norms spread roughly evenly across layers \citep{lindsey2024crosscoders, dumas2025overcoming}, we show that the functional dependence of each latent in fact collapses onto one or two on average. To measure this gap, we introduce a \emph{coherence} function, defined as the ratio of the sum of per-layer dependence to its maximum per-layer dependence (see \Cref{sec:coherence}), bounded between $1$ (all dependence on one layer) and $L$ (equal across all $L$ layers). Two choices of defining this per-layer dependence give two metrics: \emph{norm coherence}, which uses the decoder weight norm at each layer (a generalization of typical crosscoder diagnostics \citep{lindsey2024crosscoders, dumas2025overcoming}), and \emph{functional coherence}, which zeros all activations of a specific layer at the crosscoder's input and measures the resulting change in the latent's activation. We validate that, as already known in the literature \citep{lindsey2024crosscoders, dumas2025overcoming}, norm coherence is high ($>7$), but this is misleading since functional coherence is low ($\sim 2$) and thus crosscoder latents model information from just two layers on average.

This divergence reflects a structural failure of the standard crosscoder to learn cross-layer features which we trace to two concrete limitations:

\limone{Unconstrained cross-layer structure.} The encoder and decoder are parameterized as $L$ independent per-layer matrices with no structural link across layers or across the latent population. The reconstruction loss (\Cref{eq:recon_loss}) decomposes additively across layers and, as a result, the objective is invariant to reallocations of a latent's weight mass that preserve per-layer reconstructions.

\limtwo{Unregularized cross-layer dependence.} Even with a constrained structure, the reconstruction objective places no penalty on how a latent's activation depends on the different layers, so a latent can still collapse its functional dependence onto a single dominant layer at no cost.

We address these limitations by introducing \acronym, with two complementary modifications. First, \acronym constrain the cross-layer structure by stacking the per-layer decoder/encoder matrices into a 3D tensor and factorizing it~\citep{panagakis2021tensor}, directly addressing \Lone. The layer axis of the input passes through a shared low-dimensional space rather than $L$ independent matrices, so each latent's engagement across layers is drawn from a structured family shared across all latents. Second, \acronym regularize the cross-layer dependence with \emph{stochastic layer masking}, directly addressing \Ltwo. During training, the activations from specific layers are randomly masked with a probability $p$, encouraging redundant codes that distribute features across channels, similar to coding theory~\citep{marshall2024understanding,srivastava2014dropout}. In this setting, latents supported by a single layer pay a reconstruction cost growing with $p$, while latents distributed across layers do not, so the optimization pressure favors cross-layer support.

Our contributions are summarized below:

\textbf{C1.} By introducing a functional coherence metric we show that \textbf{standard crosscoders concentrate
functional dependence on two layers on average}, despite spreading their weight norms uniformly. This pathology is invisible to the norm-based diagnostics used in prior works.

\textbf{C2.} We identify two architectural limitations as the sources of the pathology described in \textbf{C1}, \textbf{unconstrained cross-layer structure} (\Lone) and \textbf{unregularized cross-layer dependence} (\Ltwo).

\textbf{C3.} To address these limitations we propose \acronym, a crosscoder variant combining two complementary modifications: (i) \textbf{low-rank tensor decomposition} of the encoder and decoder weights, which couple per-layer parameters through a shared cross-layer basis, and (ii) \textbf{stochastic layer masking}, which acts as a regularizer along the layer axis and pressures latents to spread their support across multiple layers.
    
\textbf{C4.} Across GPT2-Small, Pythia-410M, Pythia-1.4B, and Gemma2-2B, the combined method \textbf{raises probing F1} above the per-layer SAE baseline that the standard crosscoder fails to reach by large margins, \textbf{reduces reconstruction MSE} by 25--50\%, and roughly \textbf{doubles average functional coherence}. 

\textbf{C5.} A qualitative inspection together with an LLM-as-a-judge evaluation shows that \acronym\ \textbf{recover $3$--$13\times$ more semantically coherent latents} than the standard crosscoders across all LLMs, with high-coherence latents acting as concept detectors (US states, negative-affect words, etc.) and low-coherence latents collapsing onto surface-level patterns (digits, punctuation, etc.).

\section{Related Work}
\label{sec:related-work}

\textbf{Sparse dictionary learning for interpretability.}
Sparse autoencoders have become the standard tool for decomposing transformer activations into monosemantic features~\citep{bricken2023towards, cunningham2023sparse, templeton2024scaling}, with variants such as TopK~\citep{gao2024topk}, JumpReLU~\citep{rajamanoharan2024jumprelu}, and BatchTopK~\citep{bussmann2024batchtopk} refining the sparsification operation. SAEs learn a feature dictionary at a single location in the transformer, and training multiple SAEs across different transformer layers forms the basis of circuit discovery~\citep{conmy2023towards,hanna2023does,marks2024sparse}. 

\textbf{Multi-layer and cross-layer methods.}
Several approaches extend dictionary learning across layers. Multi-layer SAEs~\citep{lawson2025mlsae} and feature-matching procedures~\citep{balagansky2025permutability} align per-layer dictionaries post hoc, while crosscoders~\citep{lindsey2024crosscoders} train a single shared dictionary over all layers and provide the basis for our work. They have been used to study feature evolution within a single LLM~\citep{ameisen2025circuit, troitskii2025internal, wadell2025foundation}, and to track features across base-finetune LLM pairs or training checkpoints~\citep{baek2025towards, ge2026evolution, bayazit2025crosscoding}. The crosscoder encoders and decoders remain unconstrained, which~\citep{dumas2025overcoming} flagged in the model-axis setting as producing artifactual latents. We identify the layer-axis analogue of this failure, show it persists despite high dictionary norms across the layers, and address it by applying tensor factorization and layer masking.

\textbf{Tensor factorization in deep networks.}
Tensor decompositions including Canonical Polyadic~\citep{hitchcock1927expression} (CP) and Tensor Ring~\citep{zhao2016tensorring} (TR) have a long history of compressing weight tensors and inducing structured priors in deep models~\citep{panagakis2021tensor}. In interpretability research, tensor decompositions have been applied on the mixture-of-experts paradigm to provide scalable expert specialization~\citep{oldfield2024multilinear} and faithful approximation to multilayer perceptron (MLP) components~\citep{oldfield2025towards}, as well as to model quadratic and cubic feature interactions in polynomial SAEs~\citep{koromilas2026polysae}. In the present study, tensor decompositions are repurposed to provide a shared representation space across the latent population and across layers, addressing the structural limitation identified in standard crosscoders.

\textbf{Denoising and structured masking.}
Denoising autoencoders~\citep{vincent2008extracting, vincent2010stacked} corrupt inputs and require reconstruction of clean targets to learn robust and distributed representations; an argument that was formalized via coding theory~\citep{marshall2024understanding}. The result is redundancy in information encoding  so that masking some still leaves recoverable signal. The principle has scaled to masked language modeling~\citep{devlin2019bert}, masked image modeling~\citep{he2022masked}, and feature-level dropout~\citep{srivastava2014dropout}, all of which corrupt at the input or feature level. In our work, we apply the same principles along the layer axis where the corruption pattern correlates with the failure mode being targeted, namely layer-specialized latents.

\section{Factorized Masked Crosscoders}\label{sec:method}

\subsection{Preliminaries}
\label{sec:meth:preliminaries}

\textbf{Notation.}
Vectors and matrices are denoted by lowercase and uppercase bold letters respectively, $\vu$, $\mU$ and tensors are represented by uppercase bold upright letters $\tU$.
An element (scalar) of a matrix or tensor is accessed by subscript, \eg $\tU_{i,j,k}$.
Mode-$n$ fibers (the generalization of rows and columns from matrices to tensors) are obtained by fixing all indices except the one that corresponds to the $n^{\mathrm{th}}$ mode, \eg a mode-1 fiber of $\tU$ is $\tU_{:,j,k}$. Similarly, tensor slices are obtained by fixing a single index, \eg $\tU_{i,:,:}$.
The $i$-th column of $\mM$ is $\vm_i$, and $\mM_{:,1:r}$ denotes its first $r$ columns.
For depth stacking, \ie combining matrices as slices of a tensor along a new dimension, we use $[\mX, \mY]\in\mathbb{R}^{d_1\times d_2\times 2}$ for $X,Y\in\mathbb{R}^{d_1\times d_2}$. 
The symbol $\circ$ denotes the vector outer product, so $\va \circ \vb \circ \vc$ is the rank-one tensor with entries $a_i b_j c_k$.
$\sR^d$ and $\sR^{d_{\mathrm{sae}}}$ denote the activation and sparse-code spaces, with $d_{\mathrm{sae}} \gg d$.

\textbf{Sparse Autoencoders.}
SAEs build on overcomplete dictionary learning~\citep{mallat1993matching, gao2024topk} to decompose neural activations into a sparse set of latent features.
Given activations $\vx \in \sR^d$ from an intermediate layer of a pretrained network, an SAE learns a sparse code $\vz \in \sR^{d_{\mathrm{sae}}}$ via
$\hat{\vx} = \vb^{\mathrm{dec}} + \mD\vz,\; \vz = \cS\!\big(\mathrm{ReLU}(\mE^\top\vx + \vb^{\mathrm{enc}})\big)$,
where $\mE \in \sR^{d \times d_{\mathrm{sae}}}$ is a linear encoder, $\mD \in \sR^{d \times d_{\mathrm{sae}}}$ is the decoder dictionary, $\vb^{\mathrm{enc}}\in \sR^{d_{\mathrm{sae}}}$ and $\vb^{\mathrm{dec}}\in \sR^{d}$ are the encoder and decoder bias terms, $\hat{\vx}\in \sR^d$ are the reconstructed activations, and $\cS$ is a sparsification operator such as Top$K$~\citep{gao2024topk} or BatchTop$K$~\citep{bussmann2024batchtopk}.
Feature $i$ activation depends on the extent to which $\vx$ aligns with the encoder direction $\ve_i$ and, in the linear regime, contributes by $z_i \vd_i$ to the reconstruction.

\textbf{Crosscoders.}
A crosscoder~\citep{lindsey2024crosscoders} applies sparse dictionary learning to jointly encode activations from \emph{multiple} layers into a shared sparse latent space and use it to reconstruct all layers.
For a given input token, let $\vx_\ell \in \sR^d$ denote the activation at layer $\ell \in [1,L]$, where $L$ is the number of transformer layers considered.
The crosscoder assigns encoders $\mE_\ell \in \sR^{d \times d_{\mathrm{sae}}}$ and decoders $\mD_\ell \in \sR^{d \times d_{\mathrm{sae}}}$ to each layer to reconstruct the activations of each layer by computing, 

\begin{equation}
\vz = \cS\!\left(\mathrm{ReLU}\!\left(\sum_{\ell \in [1,L]} \mE_\ell^\top \vx_\ell + \vb^{\mathrm{enc}}\right)\right),
\qquad
\hat{\vx}_\ell = \mD_\ell\, \vz + \vb^{\mathrm{dec}}_\ell,
\label{eq:crosscoder}
\end{equation}

where $\vb^{\mathrm{enc}} \in \sR^{d_{\mathrm{sae}}}$ is a pre-activation bias shared across layers, $\vb^{\mathrm{dec}}_\ell \in \sR^{d}$ is a per-layer decoder bias, $\cS$ is a sparsification operator (e.g. Top$K$ or BatchTop$K$), and $\hat{\vx}_\ell$ is the reconstruction at layer $\ell$ produced by the encoder--decoder map. The crosscoder is trained to minimize the expected mean squared reconstruction error across all layers,
\begin{equation}
\mathcal{L}_{\mathrm{recon}} \;=\; \frac{1}{L}\sum_{\ell \in [1,L]}\,
\mathbb{E}_{\vx \sim \mathcal{D}}\!\left[\bigl\lVert \vx_\ell - \hat{\vx}_\ell(\vx) \bigr\rVert_2^2\right],
\label{eq:recon_loss}
\end{equation}
where $\vx = (\vx_1, \dots, \vx_L)$ is the tuple of per-layer activations for a token, and $\mathcal{D}$ is the joint distribution over such tuples induced by passing tokens from the training corpus through the base LLM. In practice the expectation is approximated by the empirical mean over a minibatch of tokens.

\subsection{Tensor Factorization of Crosscoder Weights}
\label{sec:meth:tensor_factorization}

Depth-stacking the per-layer encoders expresses the total encoder as a tensor $\tE = [\mE_1, \ldots, \mE_L] \in \sR^{d \times d_{\mathrm{sae}} \times L}$, and the decoder as $\tD = [\mD_1, \ldots, \mD_L] \in \sR^{d \times d_{\mathrm{sae}} \times L}$. In this setting, the mode-1 fiber $\tE_{:,i,\ell} \in \sR^{d}$ is the encoder direction for feature $i$ at layer $\ell$, and $\tD_{:,i,\ell}$ its decoder atom.

This tensor view makes the standard crosscoder's structural assumption explicit: the $L$ different frontal slices $\tE_{:,:,\ell}$ are parameterized independently, with no mechanism coupling a feature's behavior across layers. This independence is precisely the unconstrained cross-layer structure flagged as \Lone in \Cref{sec:intro}, and it manifests in two ways. First, \emph{representational fragmentation}: together with a reconstruction loss (\Cref{eq:recon_loss}) that scores each layer separately, nothing pushes latent $i$ to represent the same feature at layer $\ell$ and layer $\ell'$, so the shared latent space does not yield global features but only places layer-local features in a common coordinate system. Second, \emph{parameter redundancy}: at $\gO(d_{\mathrm{sae}} \cdot d \cdot L)$ parameters, the crosscoder costs as much as $L$ stacked SAEs and pays to re-discover cross-layer features at every layer.

We observe that both issues admit a shared fix. Inducing structure across layers resolves the fragmentation, and constraining how the three modes interact eliminates the redundancy. This choice has both empirical and theoretical precedent. Empirically, structural intervention has been shown to resolve a different crosscoder pathology, artifactual latents along the model axis in model-diffing settings~\citep{dumas2025overcoming} between a base and a fine-tuned model. While the underlying inductive bias of low-rank coupling is shared, the layer-axis failure we identify is distinct: it concerns functional dependence collapsing within a single model with many layers rather than spurious latents arising between two models, and it requires complementary regularization (\Cref{sec:meth:layer_masking}) that does not apply in the model-axis case.
Theoretically, low-rank tensor factorizations are a standard inductive bias for multi-task settings with shared latent structure across tasks~\citep{romera2013multilinear, yang2016deep}: independent per-task parameterization fails to exploit this structure even when it is present, while factorizing along the task axis biases learning toward recovering the shared representation. In our setting, each layer must reconstruct its own activations from a representation shared with the other layers, making this inductive bias directly applicable.

Concretely, we replace $\tE$ and $\tD$ with low-rank tensor decompositions that jointly factorize all three modes, forcing the input through a low-dimensional bottleneck. The main results in this study are presented using the TR decomposition because it is suited for tensors with large differences between dimensions. We also report results using the CP decomposition in Appendix~\ref{app:CP}.

\textbf{Tensor Ring Decomposition.}
The Tensor Ring (TR) decomposition~\citep{zhao2016tensorring} represents each element of the tensor as the trace of a product of matrix slices drawn from three factor tensors:
\begin{equation}
\tE_{j,i,\ell} \;=\; \Tr\!\big(\tG^{(1)}_{:,j,:}\, \tG^{(2)}_{:,i,:}\, \tG^{(3)}_{:,\ell,:}\big),
\label{eq:tr}
\end{equation}
where $\tG^{(1)} \in \sR^{R_1 \times d \times R_2}$, $\tG^{(2)} \in \sR^{R_2 \times d_{\mathrm{sae}} \times R_3}$, $\tG^{(3)} \in \sR^{R_3 \times L \times R_1}$.

TR couples the three modes through three independent dimensions, the decomposition ranks $(R_1, R_2, R_3)$, one per mode boundary in the ring. Each mode contributes a matrix and these matrices compose multiplicatively: the contribution of layer $\ell$ to $\tE_{j,i,\ell}$ is also affected by the ``feature'' slice $\tG^{(2)}_{:,i,:} \in \sR^{R_2 \times R_3}$, so the same layer pattern can act differently depending on which feature it represents. The inductive bias is therefore \emph{context-dependent coupling}: features are no longer confined to a single shared basis of layer patterns but participate in subspaces that vary with feature identity, and decomposition ranks $(R_1, R_2, R_3)$ control expressiveness separately at each mode boundary. Tensor factorization restricts the cross-layer parameterizations of $\tE$ and $\tD$ (addressing \Lone) without biasing the optimizer toward solutions that spread features across layers (\Ltwo).

\subsection{Stochastic Layer Masking}
\label{sec:meth:layer_masking}

Within the constrained encoder/decoder parametrizations, a latent can still concentrate all its functional dependence at a single layer, as nothing in the reconstruction objective of \Cref{eq:recon_loss} penalizes such a configuration. To bias learning toward the cross-layer regime, we complement the architectural constraint with an input-level perturbation, \emph{stochastic layer masking}.

For each input token and each layer $\ell$, we independently draw a Bernoulli mask $m_\ell \in \{0,1\}$ with $\Pr(m_\ell = 0) = p$, and the encoder receives the corrupted activations $\tilde{\vx}_\ell \;=\; m_\ell \cdot \vx_\ell$. The reconstruction loss targets the unmasked activations $\vx_\ell$, meaning that the crosscoder is trained to reconstruct every layer, including the masked ones. This places our objective in the denoising autoencoder framework~\citep{vincent2008extracting, vincent2010stacked} (corrupting inputs and require reconstruction of clean targets), instantiated with layer-structured masking noise, a setting in which input noise acts as an implicit Tikhonov regularizer on the encoder-decoder mapping~\citep{bishop1995training}. The masking probability $p$ directly controls how strongly the regularizer suppresses localized features. Larger $p$ makes the cross-layer pressure dominate while smaller $p$ lets localized features survive. 

A complementary perspective from coding theory shows that information traversing a sequence of channels retains capacity only when the code is redundant, and that networks trained under stochastic neuron dropout learn precisely such codes, distributing each feature across multiple units rather than localizing it~\citep{marshall2024understanding}. Stochastic layer masking applies the same principle along the layer axis: a single-layer latent pays a reconstruction cost that grows with $p$, mirroring the redundancy induced by unit-level dropout~\citep{srivastava2014dropout}. This contrasts with the windowed ``convolutional'' crosscoder of~\citep{lindsey2024crosscoders}, which architecturally restricts each latent to $K<L$ contiguous layers, imposing a hard locality constraint, opposite in effect to our stochastic regularizer.

\section{Experiments}\label{sec:experiments}

\subsection{Experimental Setup}
\label{sec:main_setup}
The training procedure for \acronym was implemented in \texttt{SAELens}~\citep{bloom2024saetrainingcodebase}, with all implementation details provided in Appendix~\ref{app:implementation}. The results presented in this study refer to crosscoders with $16{,}384$ latents trained on four pretrained base LLMs of different scales: GPT2-Small~\citep{radford2019gpt2}, Pythia-410M and Pythia-1.4B~\citep{biderman2023pythia}, and  Gemma2-2B~\citep{gemma2report}. Activations were collected from the residual stream at eight post-MLP locations spanning the middle layers of each base LLM. All crosscoders were trained using a ReLU pre-activation, followed by a BatchTop$K$~\citep{bussmann2024batchtopk} sparsifier with $K=64$. Gemma2-2B and GPT2-Small activations were generated using OpenWebText~\citep{gokaslan2019openwebtext}, while the activations from the Pythia models were generated using an uncopyrighted variant of the deduplicated Pile~\citep{gao2021pile}. Training runs used 500M tokens (300M for GPT2-Small) with a context length of 128. The evaluations that follow compare \acronym with and without layer masking against base SAEs and standard crosscoders trained with the same hyperparameters. In all main-text experiments, \acronym are matched to the standard crosscoder's parameter count by appropriate choice of decomposition ranks. Rank and masking probability ablations are presented in Appendix~\ref{app:ablations}.

\subsection{Coherence Diagnostics}\label{sec:coherence}

To assess whether a crosscoder latent neuron represents a genuinely multi-layer feature or merely responds to information from a single dominant layer, we introduce two complementary coherence metrics: \emph{norm coherence} and \emph{functional coherence}.

\paragraph{Norm Coherence.} The simplest approach to measuring layer distribution is to examine the crosscoder's weights directly, which is a generalized version of the \textit{relative decoder norms} shown in~\citep{lindsey2024crosscoders}, suitable for any number of layers. For a crosscoder with decoder weights $\tD \in \mathbb{R}^{d \times d_{\text{sae}} \times L}$, we compute the L2 norm for each latent's dictionary entry corresponding to each layer $\ell \in [1,L]$ via the decoder's mode-1 fibers $\mN_{i,\ell} = \|\tD_{:,i,\ell}\|_2$. The \emph{norm coherence} of latent $i$ is then
\begin{equation}
c_i^{\text{n}} = \frac{\sum_{\ell \in [1,L]} \mN_{i,\ell}}{\max_{\ell \in [1,L]}\mN_{i,\ell}},
\label{eq:norm_coherence}
\end{equation}
which ranges from 1 (all weight mass concentrated in one layer) to $L$ (equal norms across all layers). 

\paragraph{Functional Coherence.}
A more informative measure examines functional dependence, focusing on how a latent's activation changes when information from a specific layer is removed. For an input $\vx = (\vx_1, \ldots, \vx_L) \sim \mathcal{D}$, let $M_\ell$ denote the operator that zeros the $\ell$-th component of $\vx$, $M_\ell(\vx) = (\vx_1, \ldots, \vx_{\ell-1}, \mathbf{0}, \vx_{\ell+1}, \ldots, \vx_L)$, so that $z_i(M_\ell(\vx))$ is the latent activation when layer $\ell$ is masked while all other layers remain intact. We define latent $i$'s functional coherence $c_i^{\mathrm{f}}$ as
\begin{equation}
    c_i^{\mathrm{f}} \;=\;
        \frac{\sum_{\ell \in [1,L]} \mS_i^{\ell}}
             {\max_{\ell \in [1,L]} \mS_i^{\ell}},
    \qquad \mathrm{where} \quad
    \mS_i^{\ell} \;=\; \mathbb{E}_{\vx \sim \mathcal{D}}\!\left[
        \left|\frac{z_i(\vx) - z_i(M_\ell(\vx))}
             {z_i(\vx) + \epsilon}\right|
    \right].
    \label{eq:func_coh}
\end{equation}
$\mS_i^{\ell}$ represents the relative change of latent $i$'s activation with respect to masking layer $\ell$, with $0<\epsilon \ll 1$ a small constant for numerical stability. A high value of $\mS_i^{\ell}$ indicates that latent $i$ is sensitive to information from layer $\ell$, while a low value indicates that layer $\ell$ contributes little to that latent's activation. $c_i^{\mathrm{f}}$ therefore, measures the ratio of the total relative change across layers to the largest single-layer relative change. Analogously to norm coherence, $c_i^{\mathrm{f}}$ ranges from $1$ to $L$.

\begin{figure}[htbp]
    \centering
    \includegraphics[width=\textwidth]{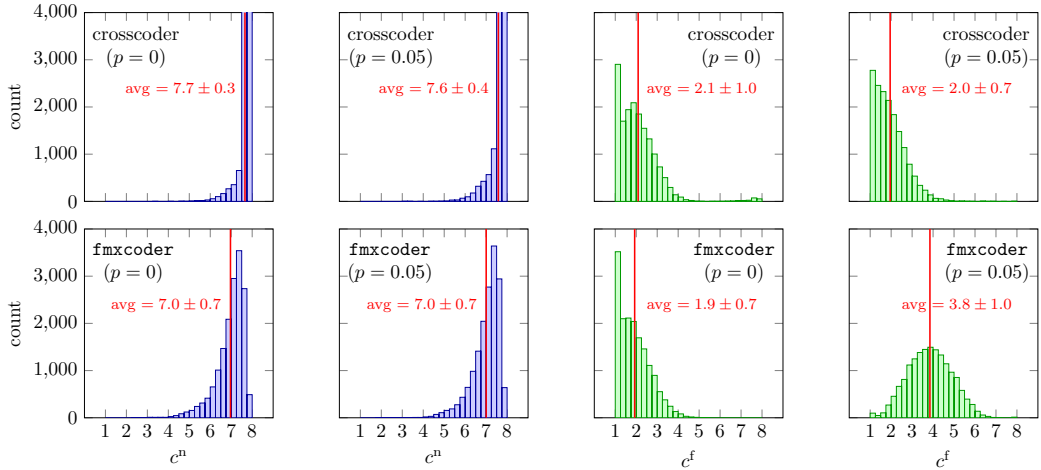}
    \caption{{\color{blue}Norm coherence $c^{\mathrm{n}}$ (blue)} and {\color{green!80!black}functional coherence $c^{\mathrm{f}}$ (green)} latent distributions for crosscoders (top row) and \acronymsingle (bottom row) variants trained on Pythia-410M. The average value and standard deviation of each coherence distribution are also shown in red color.}
    \label{fig:coherence}
\end{figure}

\Cref{fig:coherence} shows the latent distributions of norm and functional coherence for crosscoder and \acronymsingle variants trained on Pythia-410M. Norm coherence remains high (avg.\ $c^{\mathrm{n}}>7$) in all cases, while functional coherence is much lower (avg.\ $c^{\mathrm{f}}\approx2$), except for \acronymsingle with $p$=0.05, where it doubles. Factorization alone therefore does not increase functional coherence; \emph{the gain emerges only when factorization and masking are combined}. This interaction reflects the complementary roles of the two interventions. In a standard (unfactorized) crosscoder, the per-layer dictionary entries (encoder/decoder mode-1 fibers) of each latent are independent parameters. Layer masking only adjusts the weights of layers a feature already occupies, with no mechanism to extend its support to new ones, despite the correlations between residual-stream activations across layers. Tensor factorization removes this barrier by tying the per-layer dictionary entries through shared factors, so weight updates propagate jointly across layers and latents. Layer masking, applied on top, supplies the optimization pressure that turns this coupling into genuinely cross-layer features.

\subsection{Qualitative and LLM-Judge Analysis of Features}\label{sec:feature_analysis}

\Cref{tab:feature_vis} contrasts examples of highest- and lowest-$c^{\mathrm{f}}$ latents recovered by an \acronymsingle\ with $p$=0.05 trained on the activations of the Gemma2-2B model. \emph{High-$c^{\mathrm{f}}$ latents act as concept detectors} that fire on diverse tokens unified by meaning, including morphological variants and named-entity instances. \emph{Low-$c^{\mathrm{f}}$ latents instead collapse onto character classes}, single punctuation symbols, or short subword fragments with no shared semantics. Across both halves of the table, $c^{\mathrm{n}}$ remains in the narrow band 5.5--7.4 while $c^{\mathrm{f}}$ spans the full 1.0--7.1 range, confirming that weight magnitudes spread evenly across layers even when functional dependence does not, and motivating $c^{\mathrm{f}}$ as a more informative diagnostic.

\begin{table}[]
\caption{Examples of high functional coherence $c^{\mathrm{f}}$ latents (top half of the table) and low $c^{\mathrm{f}}$ latents (bottom half of the table). The corresponding norm coherence $c^{\mathrm{f}}$ is also reported. The results correspond to an \acronymsingle with $p$=0.05 trained on the activations of Gemma2-2B.}
\label{tab:feature_vis}
\centering
\footnotesize
\renewcommand{\arraystretch}{1.15}
\begin{tabular}{c l c c}
\toprule
Latent firing pattern & Description & $c^{\mathrm{f}}$ & $c^{\mathrm{n}}$  \\
\midrule
{[\textit{lobbying, campaigned, pleaded, pitch, \ldots}]}   & verbs of advocacy & 7.1  & 6.1 \\
{[\textit{regret, despise, failure, gruesome, \ldots}]}   & negative-affect words & 7.0  & 6.9 \\
{[\textit{Philadelphia, Pennsylvania, Maryland, Florida, \ldots}]}   & US states and cities & 6.9  & 6.7 \\
{[\textit{colleagues, friends, accomplice, partner, \ldots}]}   & companion relationship & 6.8  & 6.3 \\
{[\textit{vampire, werewolf, supernatural, superpowers, \ldots}]}   & supernatural beings/powers & 6.7  & 7.4 \\
\midrule
{[\textit{1, 2, 3, 4, \ldots}]}   & digit detector & 1.0  & 5.8 \\
{[\textit{def, set, define, ->, \ldots}]} & code/markup punctuation & 1.0  & 6.2 \\
{[\textit{d\#, 3, :\#, bbbb, \ldots}]} & hex/colour-code characters & 1.0  & 6.0 \\
{[\textit{G, TT, BB, QL, \ldots}]} & short sub-word fragments  & 1.0  & 6.5 \\
{[\textit{), ;, \}, ), \ldots}]} & closing-brace family  & 1.0  & 5.5 \\
\bottomrule
\end{tabular}
\end{table}

To scale this analysis, we perform an \textit{LLM-as-a-judge} evaluation, asking GPT-4o-mini as the LLM judge to score each latent on two independent axes: (i) the latent has discovered a semantic concept (firings share a meaning, concept, topic, sentiment, entity, etc.), and (ii) a surface-level pattern (grammatical role, character class, formatting, etc.). The latents are then labeled accordingly as ``semantic'' or ``surface''. Full procedure and prompts are provided in Appendix~\ref{app:llm-judge}. The total number of semantic and surface latents for all base LLMs is presented in \Cref{tab:llm-judge}, comparing \acronymsingle\ ($p$=0.05) results against crosscoder results. Across all four base LLMs, \acronymsingle\ \emph{recovers substantially more semantic latents} than the standard crosscoder, lifting the absolute count from at most a few hundred to between $1.2$k and $3.3$k. The effect is most pronounced on Gemma2-2B, the only setting in which semantic latents approach parity with surface ones. Surface-latent counts also rise on three of the four base LLMs and fall only on Gemma2-2B, so the recovered dictionary shifts toward semantic latents in composition, while the total number of clearly categorized latents increases overall.

\begin{table}[ht]
\centering
\caption{Number of semantic vs.\ surface latent counts per base LLM, using an LLM-as-a-judge procedure. Full procedure and prompts are provided in Appendix~\ref{app:llm-judge}.}
\label{tab:llm-judge}
\setlength{\tabcolsep}{6pt}
\renewcommand{\arraystretch}{1.15}
\begin{tabular}{l cc cc cc cc}
\toprule
 & \multicolumn{2}{c}{\textbf{GPT2-Small}} & \multicolumn{2}{c}{\textbf{Pythia-410M}} & \multicolumn{2}{c}{\textbf{Pythia-1.4B}} & \multicolumn{2}{c}{\textbf{Gemma2-2B}} \\
\cmidrule(lr){2-3} \cmidrule(lr){4-5} \cmidrule(lr){6-7} \cmidrule(lr){8-9}
\textbf{Method} & Sem. & Sur. & Sem. & Sur. & Sem. & Sur. & Sem. & Sur. \\
\midrule
Crosscoder & 160 & 2948 & 128 & 3184 & 399 & 8693 & 405 & 6624 \\
\acronymsingle  ($p=0.05$) & 2045 & 5934 & 1324 & 9074 & 1195 & 8925 & 3319 & 4153 \\
\bottomrule
\end{tabular}
\end{table}

\subsection{Reconstruction and Sparse Probing Results}
\label{sec:reconstruction}
For the evaluation of the trained models we report reconstruction metrics on held-out data following the SAEBench library~\citep{karvonen2025saebench}. \Cref{tab:reconstruction_evals_TR} reports the mean squared error (MSE), explained variance (EV), and cosine similarity (CS) across LLMs. We report all three metrics because they capture complementary aspects of reconstruction quality: MSE gives absolute error in native units, EV normalizes the error to enable cross-LLM comparison, and CS isolates directional fidelity, which downstream layers depend on. 
The standard crosscoder's high EV (0.986) but low CS (0.870) on Pythia-410M illustrate why all three metrics are needed. Across all four base LLMs, the standard crosscoder is the weakest reconstructor on every metric, with MSE up to roughly 100\% higher than the per-layer SAE baseline and consistent drops in CS. The \acronymsingle variants reverse this pattern, reducing MSE by 25--50\% relative to the standard crosscoder, recovering CS that, in some cases, exceeds the SAE baseline, and matching the SAE explained variance. The masking variant pays a small reconstruction cost relative to the unmasked variant, which is the expected trade-off of a denoising objective and is offset by the functional coherence gains reported in~\Cref{sec:coherence}. 

\begin{table*}[h]
\caption{Reconstruction metrics and probing-based results for different base LLMs, for SAE,  crosscoder, and \acronym variants. We report average F1 across six classification tasks and the detailed results are presented in Appendix~\ref{app:probing}. SAE results were reproduced using the settings from~\citep{koromilas2026polysae}.}
\label{tab:reconstruction_evals_TR}
\centering
\scriptsize
\renewcommand{\arraystretch}{0.9} 
\setlength{\tabcolsep}{4.5pt}
\renewcommand{\arraystretch}{1.15}
\begin{tabular}{llccc|cc}
\toprule
LLM & Variant & MSE $\downarrow$ & EV $\uparrow$& CS $\uparrow$& Avg.~F1 $\uparrow$& \makecell{Avg.~Wass.\ $\uparrow$\\$(\times 10^{-3})$}\\
 \midrule
 \multirow{4}{*}{GPT2-Small} &\ SAE & 0.53 &  0.9256 & 0.9623 & 65.7 & 8.4\\
 &\ Crosscoder & 0.64 & 0.8704 & 0.9346 & 67.6 & 11.6 \\
  &\ \acronymsingle ($p=0$) & \textbf{0.32} & \textbf{0.9359} & \textbf{0.9682} & \textbf{78.3} & \textbf{120.2}\\
 &\ \acronymsingle ($p=0.05$) & 0.33 & 0.9332 & 0.9670 & 73.5 & 68.4 \\
 \midrule
 \multirow{4}{*}{Pythia-410m} &\ SAE & \textbf{0.03} & \textbf{0.9918} & \textbf{0.9364} & 65.0 & 0.5\\
 &\ Crosscoder & 0.05 & 0.9856 & 0.8696 & 47.4 & <0.1 \\
 &\ \acronymsingle ($p=0$) & \textbf{0.03} & 0.9917 & 0.9275 & \textbf{74.3} & \textbf{10.2} \\
 &\ \acronymsingle ($p=0.05$) & \textbf{0.03} & 0.9914 & 0.9257 & 68.1 & 4.5 \\
 \midrule
 \multirow{4}{*}{Pythia-1.4b} &\ SAE &  \textbf{0.22} &  \textbf{0.9784} &  \textbf{0.9289} & 64.6 & 0.8 \\
 &\ Crosscoder & 0.31 & 0.9691 & 0.8976 & 47.0 & 0.2 \\
 &\ \acronymsingle ($p=0$) & \textbf{0.22} & 0.9780 & 0.9273 & \textbf{78.4} & \textbf{36.1}\\
 &\ \acronymsingle ($p=0.05$) & 0.23 & 0.9774 & 0.9252 & 71.2 & 13.1 \\
\midrule
 \multirow{4}{*}{Gemma2-2b} &\ SAE &  \textbf{1.58} &  \textbf{0.8702} &  0.9215 & 64.8 & 2.7\\
 &\ Crosscoder & 3.29 & 0.7809 & 0.8737 & 51.2 & 0.2 \\
 &\ \acronymsingle ($p=0$) & 2.03 & 0.8632 & \textbf{0.9229} & \textbf{81.9} &  \textbf{198.7} \\
 &\ \acronymsingle ($p=0.05$) & 2.07 & 0.8608 & 0.9218 & 73.0 & 39.1 \\
\bottomrule
\end{tabular}
\end{table*}

Furthermore, \Cref{tab:reconstruction_evals_TR} also reports sparse probing performance on six classification tasks: Bias in Bios~\citep{dearteaga2019biasinbios}, AG News~\citep{zhang2015charcnn}, EuroParl~\citep{koehn-2005-europarl}, GitHub programming languages~\citep{codeparrot_github_code}, Amazon Sentiment, and Amazon 15~\citep{hou2024bridging}. While the F1 score measures whether a latent fires on the correct class, it is invariant to how strongly the firing distributions for positive and negative examples are separated. We complement it with the 1-Wasserstein distance between the activation distributions of the two classes, computed per latent and aggregated across the probing dataset~\citep{koromilas2026polysae}. Wasserstein distance measures the minimum ``transport cost'' to convert one distribution into the other, capturing both the gap between their means and differences in their shapes. A latent with high F1 but low Wasserstein distance discriminates the classes only marginally, whereas a latent with both high F1 and high Wasserstein distance separates the classes with a wide margin, indicating a more robust feature. Averaged across all six classification tasks, the standard crosscoder underperforms the per-layer SAE baseline by 13--18 F1 points on Pythia-410M, Pythia-1.4B, and Gemma2-2B, and shows near-zero Wasserstein distance across all four base LLMs, indicating latents that barely separate the probing classes. As with reconstruction metrics, \acronym reverse this gap entirely: without masking, F1 rises to 74--82\%, exceeding the SAE baseline by 9--17 points on every model, and Wasserstein distance increases by one to three orders of magnitude, confirming that the gains reflect well-separated activation distributions rather than thin-margin classifications. The masking variants incur a small F1 cost (4--9 points) and lower Wasserstein distance, but they nonetheless remain well above the SAE and standard crosscoder baselines on F1 and Wasserstein distance. The individual sparse probing results across all the classification datasets considered in this study are presented in Appendix~\ref{app:probing}.

\section{Limitations}\label{sec:limitations}

Several aspects of the present study limit the scope of these conclusions. First, our experiments are restricted to eight post-MLP residual-stream locations spanning the middle layers of each model. The behavior of \acronym when applied to the full depth of a network, or jointly to attention-output and MLP-output streams, remains untested, and the case $L=8$ is small relative to the depth of frontier models. Second, \acronym introduce two new hyperparameters, the masking probability $p$ and the factorization ranks, whose optimal settings appear to be model- and dataset-dependent (see Appendix~\ref{app:ablations}). Our sweeps explored only a coarse grid and we do not provide an \textit{a priori} procedure for selecting $p$ when the downstream task is unknown. Third, all main-table results use a single sparsifier (BatchTop$K$ at $K$=64) and a fixed dictionary width of $16{,}384$ latents. How \acronym interact with alternative sparsity mechanisms (JumpReLU, L1) and with significantly larger dictionary widths is left to future work. Fourth, distinguishing semantic from surface latents is itself a subtle judgment, often hinging on whether a shared grammatical role also carries shared meaning, and affordable judges such as \textit{GPT-4o-mini} are not always reliable at this distinction. Fifth, \acronym incur a slight additional training-time cost relative to standard crosscoders (<10\%) due to the tensor-factorization forward/backward passes, with compute costs reported in Appendix~\ref{app:implementation}. Finally, all main-table results come from a single training run per LLM/\acronymsingle variant due to compute constraints.

\section{Conclusion}

We introduced the \acronymsingle architecture, that uses low-rank tensor decomposition for the encoder/decoder tensors and applies stochastic layer masking to act as a remedy to standard crosscoders' tendency to learn primarily layer-localized features. Together, the two interventions raise mean probing F1 by 10--30 points across GPT2-Small, Pythia-410M, Pythia-1.4B, and Gemma2-2B, surpassing per-layer SAE baselines that the standard crosscoder fails to reach, reduce reconstruction MSE by 25--50\%, and roughly double the average functional coherence of the recovered latents. A qualitative inspection further indicates that high-coherence latents act as semantic concept detectors (US states and cities, negative-affect words, verbs of advocacy), while low-coherence latents collapse onto surface-level character classes and syntactic tokens. An LLM-as-a-judge evaluation corroborates this shift quantitatively: \acronym recover 1.2k--3.3k semantic concept latents per model, roughly an order of magnitude above the standard crosscoder, with semantic latents approaching parity with surface ones on Gemma2-2B.

\bibliographystyle{IEEEtran}
\bibliography{main}

\medskip

{
\small


\appendix

\section{Implementation Details}
\label{app:implementation}

This appendix documents the training setup, hyperparameters, and infrastructure used for all experiments in the main paper. All crosscoder variants (standard crosscoder, TR \acronymsingle, and CP \acronymsingle) were trained with identical optimizer, schedule, and data settings; the only differences are the encoder/decoder parameterization and the presence of stochastic layer masking.

\subsection{Architectures and Hyperparameters}

\textbf{Dictionary configuration.} All crosscoders and \acronym\ use a dictionary width of $d_{\mathrm{sae}} = 16{,}384$ latents. The sparsity operator is BatchTopK~\citep{bussmann2024batchtopk} with $K=64$ active latents per token. A pre-activation ReLU is applied before BatchTopK selection. The encoder bias $\vb^{\mathrm{enc}} \in \mathbb{R}^{d_{\mathrm{sae}}}$ is shared across layers; per-layer decoder biases $\vb^{\mathrm{dec}}_\ell \in \mathbb{R}^d$ are learned independently for each $\ell \in [1, L]$.

\textbf{Layer selection.} For each base LLM, we use $L=8$ residual-stream activations collected at post-MLP positions spanning the middle layers. The specific layer indices are (zero-indexed):
\begin{itemize}
    \item GPT2-Small (12 layers): layers 2--9.
    \item Pythia-410M (24 layers): layers 8--15.
    \item Pythia-1.4B (24 layers): layers 8--15.
    \item Gemma2-2B (26 layers): layers 9--16.
\end{itemize}

\textbf{Factorization ranks.} For TR \acronym matching the standard crosscoder parameter count, the ranks $(R_1, R_2, R_3)$ were chosen subject to $R_2/R_1 = \sqrt{d/L}$ and $R_3/R_1 = \sqrt{d_{\mathrm{sae}}/d}$. The specific values per base LLM are:
\begin{itemize}
    \item GPT2-Small ($d=768$): $(5, 244, 25)$.
    \item Pythia-410M ($d=1024$): $(7, 302, 27)$.
    \item Pythia-1.4B ($d=2048$): $(11, 501, 31)$.
    \item Gemma2-2B ($d=2304$): $(12, 545, 32)$.
\end{itemize}
For CP \acronym, the rank $R$ was chosen to match the standard crosscoder parameter count; specific values per base LLM are:
\begin{itemize}
    \item GPT2-Small: $R=5866$.
    \item Pythia-410M: $R=7707$.
    \item Pythia-1.4B: $R=14557$.
    \item Gemma2-2B: $R=16153$.
\end{itemize}

\textbf{Masking probability.} The default masking probability for \acronym reported in the main results is $p = 0.05$. The unmasked variant uses $p=0$. Bernoulli masks are drawn independently per token and per layer at every training step. Ablations of the masking probability are presented in Appendix~\ref{app:ablations}.

\subsection{Optimization}

\textbf{Optimizer.} We use Adam with a learning rate $3\times 10^{-4}$, $(\beta_1, \beta_2) = (0.9,0.999)$, with no warmup or decay schedules. For training stabilization, we use gradient clipping with a maximum norm of 1.0.

\textbf{Batch size.} 4096.

\textbf{Training tokens.} All training runs use 500M tokens (300M for GPT2-Small) at a context length of 128. 

\textbf{Initialization.} Encoder and decoder tensors / decomposition factors follow a variance-matched i.i.d.\ Gaussian initialization, chosen so that the materialized 3-D weight tensor has the same per-element variance as a Kaiming-uniform-initialized 2-D slice. Biases are initialized as zero.

\textbf{Decoder normalization.} No decoder normalization is applied (e.g.\ similar to~\citep{bricken2023towards}).

\subsection{Data}

\textbf{Activation generation.} For Gemma2-2B and GPT2-Small, activations were generated using OpenWebText~\citep{gokaslan2019openwebtext}. For Pythia-410M and Pythia-1.4B, activations were generated using an uncopyrighted variant of the deduplicated Pile~\citep{gao2021pile}. All activations are cached at the post-MLP residual-stream position for each of the $L=8$ selected layers.

\textbf{Held-out evaluation set.} Reconstruction metrics are computed on a held-out set of approximately 410k tokens not seen during training. The same held-out set is used for all variants of a given base LLM to ensure direct comparability.

\subsection{Sparse Probing Setup}

We follow the protocol of~\citep{koromilas2026polysae} for sparse probing. For each classification task, we select the top-$K=1$ latent that best separates classes by F1 on a training split, then report F1 and Wasserstein distance on a held-out evaluation split. Class labels and dataset splits follow the original sources for Bias in Bios~\citep{dearteaga2019biasinbios}, AG News~\citep{zhang2015charcnn}, EuroParl~\citep{koehn-2005-europarl}, GitHub programming languages~\citep{codeparrot_github_code}, Amazon Sentiment, and Amazon 15~\citep{hou2024bridging}.

\subsection{Functional Coherence Computation}

The functional coherence metric (\Cref{sec:coherence}) is computed on a held-out set of 10M tokens, in order to obtain enough sample size for individual latent evaluations. For each layer $\ell \in [1, L]$, we run an additional forward pass through the crosscoder with $\vx_\ell$ set to zero while leaving the other layers intact, and record the resulting per-latent activations. The numerical-stability constant in \Cref{eq:func_coh} is set to $\epsilon = 10^{-8}$. 

\subsection{Compute and Reproducibility}

\textbf{Hardware.} All experiments were run on NVIDIA H100 64GB  GPUs, with a single GPU per training run.

\textbf{Wall-clock time.} A single \acronymsingle run takes approximately 12 hours for GPT2-Small, 16 hours for Pythia-410M, 30 hours for Pythia-1.4B, and 42 hours for Gemma2-2B. 

\textbf{Total compute.} The full set of experiments reported in the main paper and appendices required approximately 1500 GPU hours of compute across all base LLMs, factorization variants, and ablation settings.

\textbf{Software.} Crosscoder training is implemented in SAELens~\citep{bloom2024saetrainingcodebase}, a PyTorch-based SAE training library in Python. Tensor Ring and CP factorizations are implemented as native PyTorch modules.  Reconstruction metrics (MSE, EV, CS) are computed following the protocols of SAEBench~\citep{karvonen2025saebench}. Probing F1 and Wasserstein distance follow~\citep{koromilas2026polysae}. The LLM-as-a-judge evaluation calls GPT-4o-mini through the OpenRouter API.

\section{CP \acronym}\label{app:CP}
\subsection{Mathematical definition}
The Canonical Polyadic (CP) decomposition~\citep{hitchcock1927expression,harshman1970foundations} approximates a tensor as a sum of rank-one tensors formed from vector outer products. For the encoder tensor with rank $R$,
\begin{equation}
\tE \;\approx\; \sum_{r \in [1,R]} \vw_r \circ \vu_r \circ \vv_r,
\qquad \tE_{j,i,\ell} \;\approx\; \sum_{r \in [1,R]} W_{jr}\, U_{ir}\, V_{\ell r},
\label{eq:cp_elementwise}
\end{equation}
where $\vw_r \in \sR^{d}$, $\vu_r \in \sR^{d_{\mathrm{sae}}}$, $\vv_r \in \sR^{L}$ are the factor vectors, collected into factor matrices $\mW \in \sR^{d \times R}$, $\mU \in \sR^{d_{\mathrm{sae}} \times R}$, $\mV \in \sR^{L \times R}$. The decoder factorizes independently with factors $\tilde{\mW}, \tilde{\mU}, \tilde{\mV}$ of matching shapes.

Each rank-one term is an outer product of an activation-space direction $\vw_r$, a feature-weight vector $\vu_r$, and a layer pattern $\vv_r$, making the full encoder a sum of $R$ atoms shared across all features. This results in a \emph{uniform coupling} inductive bias: a single rank $R$ ties the three modes together at once, and every feature expresses its cross-layer behavior in the same $R$-dimensional basis of layer patterns (the rows of $\mV$). Crucially, a feature that appears across layers \textit{is represented once} through its $R$-dimensional coefficients rather than re-discovered at each layer, and the parameter count drops from $\gO(L \cdot d_{\mathrm{sae}} \cdot d)$ to $\gO(R \cdot (d + d_{\mathrm{sae}} + L))$ per tensor.

\subsection{Results}
This section presents reconstruction and sparse-probing results for CP \acronym, reported in \Cref{tab:reconstruction_evals_CP}. The qualitative picture matches the TR results in \Cref{tab:reconstruction_evals_TR}: CP \acronym improve over the standard crosscoder on every metric and every LLM, reducing MSE by up to 50\%, recovering cosine similarity to within 0.01 of the per-layer SAE baseline, and lifting average probing F1 by up to 30 points above the crosscoder. As in the TR case, the masking variant trades a small amount of reconstruction quality and F1 for a denoising effect that is reflected in the functional coherence metric (see Appendix~\ref{app:ablations}). Comparing the two factorizations directly, CP \acronym match TR \acronym closely on MSE, explained variance, and cosine similarity, but under-perform on average F1 on 3 out of 4 LLMs, and on Wasserstein distance (by roughly a factor of 2--5 $\times$). They nonetheless remain well above the crosscoder baseline on Wasserstein, by one to two orders of magnitude. We attribute this gap to the structural difference between the two decompositions discussed in \Cref{sec:meth:tensor_factorization}: TR couples the three modes through three independent ranks ($R_1$, $R_2$, $R_3$) with context-dependent interaction across features and layers, while CP ties all three modes through a single shared rank $R$ and a uniform basis of layer patterns. The TR parameterization therefore admits richer per-feature layer profiles, more suitable for our setting where $L<<d_{sae}$, which appears to translate into more separable latent activations on the probing tasks.

\begin{table*}[h]
\caption{Reconstruction metrics and probing-based results for different base LLMs, for SAE,  crosscoder, and \acronymsingle variants. \acronym adopt the CP decomposition, with ranks chosen to match the parameters of the crosscoders in each case. We report average F1 across six classification tasks and the detailed results are presented in Appendix~\ref{app:probing}. SAE results were reproduced using the settings from~\citep{koromilas2026polysae}.}
\label{tab:reconstruction_evals_CP}
\centering
\scriptsize
\renewcommand{\arraystretch}{0.9} 
\setlength{\tabcolsep}{4.5pt}
\renewcommand{\arraystretch}{1.15}
\begin{tabular}{llccc|cc}
\toprule
LLM & Variant & MSE & EV $\uparrow$& CS $\uparrow$& Avg.~F1 $\uparrow$& \makecell{Avg.~Wass.\ $\uparrow$\\$(\times 10^{-3})$}\\
 \midrule
 \multirow{4}{*}{GPT2-Small} &\ SAE & 0.53 &  0.9256 & 0.9623 & 65.7 & 8.4\\
 &\ Crosscoder & 0.64 & 0.8704 & 0.9346 & 67.6 & 11.6 \\
 &\ \acronymsingle ($p=0$) & \textbf{0.32} & \textbf{0.9355} & \textbf{0.9679} & \textbf{74.3} & \textbf{25.8}\\
 &\ \acronymsingle ($p=0.05$) & 0.33 & 0.9330 & 0.9668 & 69.7 & 18.5 \\
 \midrule
 \multirow{4}{*}{Pythia-410m} &\ SAE & \textbf{0.03} & \textbf{0.9918} & \textbf{0.9364} & 65.0 & 0.5\\
 &\ Crosscoder & 0.05 & 0.9856 & 0.8696 & 47.4 & <0.1 \\
 &\ \acronymsingle ($p=0$)& \textbf{0.03} & 0.9917 & 0.9273 & \textbf{77.5} &  \textbf{4.7} \\
 &\ \acronymsingle ($p=0.05$) & \textbf{0.03} & 0.9915 & 0.9263 & 65.6 & 2.1 \\
 \midrule
 \multirow{4}{*}{Pythia-1.4b} &\ SAE &  \textbf{0.22} &  \textbf{0.9784} &  \textbf{0.9289} & 64.6 & 0.8 \\
 &\ Crosscoder & 0.31 & 0.9691 & 0.8976 & 47.0 & 0.2 \\
 &\ \acronymsingle ($p=0$)& \textbf{0.22} & 0.9781 & 0.9275 & \textbf{76.2} & \textbf{10.8} \\
 &\ \acronymsingle ($p=0.05$) & 0.23 & 0.9777 & 0.9263 & 65.8 & 5.5 \\
\midrule
 \multirow{4}{*}{Gemma2-2b} &\ SAE &  \textbf{1.58} &  \textbf{0.8702} &  0.9215 & 64.8 & 2.7\\
 &\ Crosscoder & 3.29 & 0.7809 & 0.8737 & 51.2 & 0.2 \\
 &\ \acronymsingle ($p=0$)& 2.05 & 0.8622 & \textbf{0.9226} & \textbf{79.0} & \textbf{38.3} \\
 &\ \acronymsingle ($p=0.05$) & 2.05 & 0.8621 & 0.9223 & 65.4 & 16.9 \\
\bottomrule
\end{tabular}
\end{table*}

\section{Mask Probability and Rank Ablations}\label{app:ablations}

This appendix reports ablations of the two hyperparameters introduced by \acronym: the masking probability $p$ and the factorization ranks. \Cref{fig:heatmap_TR} shows results for the TR variant and \Cref{fig:heatmap_CP} for the CP variant, both trained on Pythia-410m. Each figure reports MSE, mean probing F1, and average functional coherence $c^{\text{f}}$ as $p$ varies in $\{0, 0.02, 0.05, 0.10\}$ and the parameter count is reduced by factors of $1$, $1/2$, $1/4$, and $1/8$ relative to the standard crosscoder. The ranks were chosen to match the target parameter count subject to the constraints $R_2/R_1 = \sqrt{d/L}$ and $R_3/R_1 = \sqrt{d_{\text{sae}}/d}$, which balance the contribution of each mode to the factorization. The corresponding rank settings are listed in the figure captions.

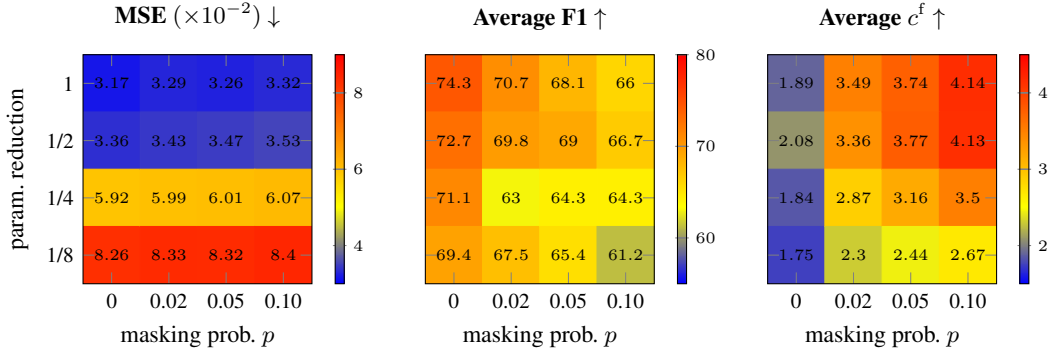
\begin{figure}[!t]
\centering
\begin{tikzpicture}
    \begin{groupplot}[
        group style={
            group size=3 by 1,
            horizontal sep=1.5cm,
        },
        width=0.33\textwidth,
        height=0.33\textwidth,
        view={0}{90},
        xlabel={masking prob.\ $p$},
        ylabel={param.\ reduction},
        tick label style={font=\footnotesize},
        label style={font=\small},
        title style={font=\bfseries\small},
        xtick={1,2,3,4},
        xticklabels={0,0.02,0.05,0.10},
        ytick={1,2,3,4},
        yticklabels={1,1/2,1/4,1/8},
        xmin=0.5, xmax=4.5,
        ymin=0.5, ymax=4.5,
        colorbar,
        colorbar style={width=0.12cm, tick label style={font=\tiny}},
        shader=flat,
    ]
    \nextgroupplot[
        title={MSE $(\times 10^{-2})$ $\downarrow$},
        point meta min=3,
        point meta max=9,
    ]
    \addplot[
        matrix plot,
        mesh/rows=4,
        mesh/cols=4,
        point meta=explicit,
        nodes near coords,
        nodes near coords style={font=\tiny, anchor=center, color=black},
    ] table [meta=v] {
        x y v
        1 1 3.17
        1 2 3.36
        1 3 5.92
        1 4 8.26
        2 1 3.29
        2 2 3.43
        2 3 5.99
        2 4 8.33
        3 1 3.26
        3 2 3.47
        3 3 6.01
        3 4 8.32
        4 1 3.32
        4 2 3.53
        4 3 6.07
        4 4 8.40
    };

    \nextgroupplot[
        title={Average F1 $\uparrow$},
        ylabel={},
        yticklabels={,,},
        point meta min=55,
        point meta max=80,
    ]
    \addplot[
        matrix plot,
        mesh/rows=4,
        mesh/cols=4,
        point meta=explicit,
        nodes near coords,
        nodes near coords style={font=\tiny, anchor=center, color=black},
    ] table [meta=v] {
        x y v
        1 1 74.3
        1 2 72.7
        1 3 71.1
        1 4 69.4
        2 1 70.7
        2 2 69.8
        2 3 63.0
        2 4 67.5
        3 1 68.1
        3 2 69.0
        3 3 64.3
        3 4 65.4
        4 1 66.0
        4 2 66.7
        4 3 64.3
        4 4 61.2
    };

    \nextgroupplot[
        title={Average $c^{\mathrm{f}}$ $\uparrow$},
        ylabel={},
        yticklabels={,,},
        point meta min=1.5,
        point meta max=4.5,
    ]
    \addplot[
        matrix plot,
        mesh/rows=4,
        mesh/cols=4,
        point meta=explicit,
        nodes near coords,
        nodes near coords style={font=\tiny, anchor=center, color=black},
    ] table [meta=v] {
        x y v
        1 1 1.89
        1 2 2.08
        1 3 1.84
        1 4 1.75
        2 1 3.49
        2 2 3.36
        2 3 2.87
        2 4 2.30
        3 1 3.74
        3 2 3.77
        3 3 3.16
        3 4 2.44
        4 1 4.14
        4 2 4.13
        4 3 3.50
        4 4 2.67
    };
    \end{groupplot}
\end{tikzpicture}
\caption{Heat maps for MSE, mean F1 and mean functional coherence $c^{\mathrm{f}}$, for different values of the masking probability $p$ and parameter reduction associated with decreasing factorization ranks for TR \acronym. $(R_1,R_2,R_3)$ take the values $(7,302,27)$ for matching the unfactorized crosscoder number of parameters, $(5,214,19)$ for a $1/2$ reduction, $(3,151,13)$ for a $1/4$ reduction, and $(2,107,9)$ for a $1/8$ reduction. In all cases, the activations were collected from Pythia-410m.}
\label{fig:heatmap_TR}
\end{figure}
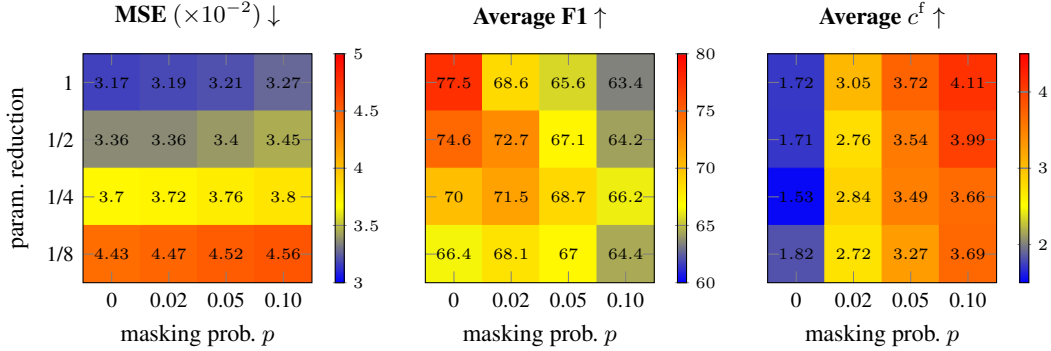
\begin{figure}[!t]
\centering
\begin{tikzpicture}
    \begin{groupplot}[
        group style={
            group size=3 by 1,
            horizontal sep=1.5cm,
        },
        width=0.33\textwidth,
        height=0.33\textwidth,
        view={0}{90},
        xlabel={masking prob.\ $p$},
        ylabel={param.\ reduction},
        tick label style={font=\footnotesize},
        label style={font=\small},
        title style={font=\bfseries\small},
        xtick={1,2,3,4},
        xticklabels={0,0.02,0.05,0.10},
        ytick={1,2,3,4},
        yticklabels={1,1/2,1/4,1/8},
        xmin=0.5, xmax=4.5,
        ymin=0.5, ymax=4.5,
        colorbar,
        colorbar style={width=0.12cm, tick label style={font=\tiny}},
        shader=flat,
    ]
    \nextgroupplot[
        title={MSE $(\times 10^{-2})$ $\downarrow$},
        point meta min=3,
        point meta max=5,
    ]
    \addplot[
        matrix plot,
        mesh/rows=4,
        mesh/cols=4,
        point meta=explicit,
        nodes near coords,
        nodes near coords style={font=\tiny, anchor=center, color=black},
    ] table [meta=v] {
        x y v
        1 1 3.17
        1 2 3.36
        1 3 3.70
        1 4 4.43
        2 1 3.19
        2 2 3.36
        2 3 3.72
        2 4 4.47
        3 1 3.21
        3 2 3.40
        3 3 3.76
        3 4 4.52
        4 1 3.27
        4 2 3.45
        4 3 3.80
        4 4 4.56
    };

    \nextgroupplot[
        title={Average F1 $\uparrow$},
        ylabel={},
        yticklabels={,,},
        point meta min=60,
        point meta max=80,
    ]
    \addplot[
        matrix plot,
        mesh/rows=4,
        mesh/cols=4,
        point meta=explicit,
        nodes near coords,
        nodes near coords style={font=\tiny, anchor=center, color=black},
    ] table [meta=v] {
        x y v
        1 1 77.5
        1 2 74.6
        1 3 70.0
        1 4 66.4
        2 1 68.6
        2 2 72.7
        2 3 71.5
        2 4 68.1
        3 1 65.6
        3 2 67.1
        3 3 68.7
        3 4 67.0
        4 1 63.4
        4 2 64.2
        4 3 66.2
        4 4 64.4
    };

    \nextgroupplot[
        title={Average $c^{\mathrm{f}}$ $\uparrow$},
        ylabel={},
        yticklabels={,,},
        point meta min=1.5,
        point meta max=4.5,
    ]
    \addplot[
        matrix plot,
        mesh/rows=4,
        mesh/cols=4,
        point meta=explicit,
        nodes near coords,
        nodes near coords style={font=\tiny, anchor=center, color=black},
    ] table [meta=v] {
        x y v
        1 1 1.72
        1 2 1.71
        1 3 1.53
        1 4 1.82
        2 1 3.05
        2 2 2.76
        2 3 2.84
        2 4 2.72
        3 1 3.72
        3 2 3.54
        3 3 3.49
        3 4 3.27
        4 1 4.11
        4 2 3.99
        4 3 3.66
        4 4 3.69
    };
    \end{groupplot}
\end{tikzpicture}
\caption{Similar to~\Cref{fig:heatmap_TR}, for CP \acronym. $R$=7707 for matching the unfactorized crosscoder number of parameters, $R$=3853 for a $1/2$ reduction, $R$=1926 for a $1/4$ reduction, and $R$=963 for a $1/8$ reduction.}
\label{fig:heatmap_CP}
\end{figure}

\paragraph{Effect of masking probability.} Across both factorizations, $p$ controls a clean trade-off between reconstruction quality and functional coherence. Functional coherence increases monotonically with $p$, roughly doubling from $c^{\text{f}} \approx 2$ at $p$=0 to $c^{\text{f}} \approx 4$ at $p$=0.1 for both TR and CP at full rank. Reconstruction MSE is essentially flat in $p$ at any fixed rank (e.g.\ $3.17 \to 3.32$ for TR at full rank), confirming that masking exerts pressure on \emph{which latents are learned} rather than on overall reconstruction capacity. Average F1 declines moderately with $p$ (up to 14 points), consistent with the denoising objective penalizing layer-localized features in favor of more distributed ones. The choice of $p$ is therefore a knob the user can set based on whether downstream interpretability or sparse-probing performance is the priority: small $p$ ($\leq 0.02$) preserves probing F1 with modest coherence gains, while $p \geq 0.05$ produces substantially more cross-layer latents at a small reconstruction and F1 cost.

\paragraph{Effect of factorization rank.} Reducing the rank reduces parameter count and degrades reconstruction and F1, but does not affect functional coherence. At $p$=0, $c^{\text{f}}$ remains in the narrow band $1.5$--$2.1$ across the full $8\times$ compression sweep, indicating that capacity controls how well features are reconstructed but not how distributed they are across layers. Functional coherence is evidently governed by the optimization pressure from masking, not by the size of the parameter budget. The two factorizations differ markedly in how gracefully they tolerate compression: CP's MSE rises only from approximately $3.2$ to $4.5$ at $1/8$ reduction, while TR's rises from approximately $3.2$ to $8.3$, more than doubling.

\paragraph{Joint behavior.} The two hyperparameters act largely independently: $p$ shifts coherence with weak effect on reconstruction, while rank shifts reconstruction and F1 with no effect on coherence. The best F1 is consistently obtained at full rank with $p$=0 ($77.5$ for CP, $74.3$ for TR), and the best coherence at full rank with $p$=0.1 ($c^{\text{f}} \approx 4.1$ for both). No setting in the sweep simultaneously maximizes both, reflecting the fundamental trade-off introduced by the denoising objective; intermediate values such as $p$=0.05 at full rank offer a reasonable balance and are the settings used in the main results.

\section{Full probing results}\label{app:probing}
\Cref{tab:full_probing} presents the per-dataset probing results underlying the averages reported in \Cref{tab:reconstruction_evals_TR} and \Cref{tab:reconstruction_evals_CP}. Three patterns are worth noting. First, the standard crosscoder's under-performance relative to the per-layer SAE is broad rather than driven by any single dataset, with losses of 10--30 F1 points across most LLM and dataset combinations on the three larger models. Second, the gains of TR \acronym are similarly systematic, beating the SAE on 22 of the 24 LLM and dataset combinations. Third, gaps in Wasserstein distances are larger than gaps in F1 in relative terms, with \acronymsingle values exceeding the crosscoder by one to three orders of magnitude even on datasets where F1 improvements are modest, supporting the main-text observation that F1 alone understates the separation gain. The comparison between TR and CP is dataset-dependent: TR achieves the higher average F1 on 3 out of 4 LLMs, but CP wins on individual datasets in several cases.

\begin{table*}[h]
\caption{Probing experiments across datasets at K=1. Format: F1 scores (\%)  / Wasserstein distance $(\times 10^{-3})$. The last two columns show average F1 and average Wasserstein distance. \acronym correspond to p=0.}
\label{tab:full_probing}
\centering
\scriptsize
\renewcommand{\arraystretch}{0.9} 
\setlength{\tabcolsep}{4.5pt}
\renewcommand{\arraystretch}{1.15}
\begin{tabular}{llcccccc|cc}
\toprule
LLM & Variant & EuroParl & Bios & \makecell{Amazon \\ Sentiment} & GitHub & AG News & Amazon 15 & Avg.~F1 (\%) $\uparrow$ & \makecell{Avg.~Wass.\ $\uparrow$\\$(\times 10^{-3})$}\\
 \midrule
 \multirow{4}{*}{GPT2-Small} &\ SAE &  67.4 / 19.0 &  59.6 / 7.3 &  68.8 / 4.4 &  68.1 / 8.8 &  65.3 / 8.2 &  65.1 / 2.9 & 65.7 & 8.4\\
 &\ Crosscoder & 85.0 / 23.0 & 55.7 / 6.2 & 67.9 / 7.5 & 66.6 / 20.4 & 62.5 / 6.0 & 68.1 / 6.2 & 67.6 & 11.6 \\
 &\ CP \acronymsingle & 84.8 / 55.7 & 72.3 / 24.3 & 81.9 / 11.3 & 68.9 / 28.0 & 66.8 / 26.8 & 71.1 / 8.5 & 74.3 & 25.8\\
  &\ TR \acronymsingle & 93.2 / 256.6 & 78.8 / 114.7 & 82.1 / 53.8 & 67.8 / 126.1 & 75.6 / 130.9 & 72.5 / 38.9 & 78.3 & 120.2\\
 \midrule
 \multirow{4}{*}{Pythia-410m} &\ SAE &  90.9 / 0.8 &  60.5 / 0.3 &  63.9 / 0.4 &  59.7 / 1.1 &  58.7 / 0.3 &  56.6 / 0.3 & 65.0 & 0.5\\
 &\ Crosscoder & 58.1 / 0.2 & 31.4 / 0.0 & 55.6 / 0.0 & 58.7 / 0.0 & 33.9 / 0.0 & 46.7 / 0.0 & 47.4 & 0.0 \\
 &\ CP \acronymsingle & 91.5 / 7.0 & 74.9 / 4.4 & 79.7 / 2.7 & 72.0 / 7.8 & 73.6 / 4.3 & 73.0 / 1.9 & 77.5 &  4.7 \\
  &\ TR \acronymsingle & 87.3 / 13.7 & 65.1 / 8.9 & 86.6 / 6.8 & 75.6 / 18.6 & 69.3 / 8.8 & 61.9 / 4.5 & 74.3 & 10.2 \\
 \midrule
 \multirow{4}{*}{Pythia-1.4b} &\ SAE &  74.0 / 0.6 &  65.0 / 0.5 &  57.2 / 0.6 &  63.3 / 2.3 &  65.2 / 0.4 &  63.2 / 0.4 & 64.6 & 0.8 \\
 &\ Crosscoder & 40.1 / 0.0 & 40.2 / 0.0 & 50.7 / 0.2 & 59.3 / 0.9 & 33.5 / 0.0 & 58.5 / 0.2 & 47.0 & 0.2 \\
 &\ CP \acronymsingle & 98.7 / 14.1 & 70.4 / 9.6 & 88.1 / 6.4 & 60.4 / 20.4 & 69.2 / 9.6 & 70.6 / 4.6 & 76.2 & 10.8 \\
  &\ TR \acronymsingle & 94.3 / 46.6 & 72.2 / 34.0 & 89.2 / 22.9 & 78.4 / 63.1 & 66.6 / 34.3 & 69.6 / 15.8 & 78.4 & 36.1\\
\midrule
 \multirow{4}{*}{Gemma2-2b} &\ SAE &  68.3 / 1.9 &  67.6 / 2.6 &  71.1 / 2.8 &  64.4 / 4.5 &  59.9 / 2.7 &  57.6 / 1.9 & 64.8 & 2.7\\
 &\ Crosscoder & 69.9 / 1.0 & 55.6 / 0.2 & 36.6 / 0.0 & 51.4 / 0.0 & 50.7 / 0.1 & 43.0 / 0.0 & 51.2 & 0.2 \\
 &\ CP \acronymsingle & 94.8 / 38.3 & 81.5 / 47.4 & 93.1 / 28.7 & 75.3 / 46.7 & 59.9 / 49.1 & 69.6 / 19.6 & 79.0 & 38.3 \\
  &\ TR \acronymsingle & 97.3 / 190.5 & 82.0 / 252.1 & 93.3 / 150.9 & 86.1 / 231.8 & 63.8 / 268.0 & 68.7 / 99.2 & 81.9 &  198.7 \\
\bottomrule
\end{tabular}
\end{table*}

\section{LLM-as-a-Judge Procedure}
\label{app:llm-judge}

For each latent we collect its top-activating tokens, sorted by maximum activation across the residual-stream activations of the base LLM on a held-out subset of its training corpus. We additionally include up to two context windows of $80$ characters on each side of the activating token. The resulting prompt is sent to \textit{GPT-4o-mini} via the OpenRouter API, with a system message instructing it to evaluate the latents. The returned \texttt{semantic\_score} and \texttt{surface\_score} are each clipped to $[0,1]$, and a latent is labeled \emph{semantic} when $\texttt{semantic\_score}>0.7$ and $\texttt{surface\_score}<0.3$, and \emph{surface} with these conditions reversed. The full user-message template is shown below:
\begin{tcolorbox}[
    colback=gray!8,
    colframe=gray!8,
    boxrule=0pt,
    arc=2pt,
    left=8pt, right=8pt, top=6pt, bottom=6pt,
    breakable
]
\begin{Verbatim}[fontsize=\footnotesize, breaklines=true, breakanywhere=true,
                 breaksymbolleft={}, breaksymbolright={}]
You are evaluating a latent feature from a sparse autoencoder trained on a language model. You are given the tokens that most strongly activate this feature.

Score the feature on two independent dimensions:

**semantic_score** (0 to 1, continuous): How strongly do these tokens collectively represent a coherent, high-level concept that is interpretable by humans?
- 1.0 = tokens clearly belong to a unified semantic category (e.g. US states and cities, negative-emotion words, cooking verbs, animal species, financial terms)
- 0.5 = tokens share some thematic connection but it's loose or partial
- 0.0 = no discernible high-level concept; tokens seem unrelated or random

**surface_score** (0 to 1, continuous): How strongly do these tokens collectively describe low-level linguistic or surface patterns rather than meaning?
- 1.0 = tokens are unified by syntax, morphology, punctuation, character class, or formatting (e.g. closing braces, digits, -ing suffixes, markup tags, subword fragments)
- 0.5 = tokens partially share surface-level properties but also carry some semantic content
- 0.0 = tokens are not unified by any surface-level pattern

These two scores are INDEPENDENT and can vary continuously from 0 to 1. A feature can be high on both (rare), low on both (noise/random), or high on one and low on the other (typical).

Think carefully about whether the unifying pattern is semantic (about meaning) or surface-level (about form/syntax/characters).

**Feature ID**: {feature_id}
**Top-activating tokens** (ordered by activation strength): {tokens}

**Example contexts** (activating token wrapped in <<>>):
- token "{tok}":
    - {context_window}
    ...

Return ONLY valid JSON with keys "semantic_score" and "surface_score", both floats between 0 and 1.
Example: {"semantic_score": 0.85, "surface_score": 0.1}
\end{Verbatim}
\end{tcolorbox}

\end{document}